\documentclass[sigconf]{acmart}
\renewcommand\footnotetextcopyrightpermission[1]{} 
\usepackage{balance}

\AtBeginDocument{%
  }

\acmConference[Preprint]{Preprint}{2025}{Sydney}

\begin{document}

\title{TrajLLM: A Modular LLM-Enhanced Agent-Based Framework for Realistic Human Trajectory Simulation}

\author{Chenlu Ju}
\authornote{All authors contributed equally to this work.}
\affiliation{%
  \institution{University of New South Wales}
  \city{Sydney}
  \state{NSW}
  \country{Australia}
}
\email{c.ju@student.unsw.edu.au}

\author{Jiaxin Liu}
\authornotemark[1]
\affiliation{%
  \institution{University of New South Wales}
  \city{Sydney}
  \state{NSW}
  \country{Australia}
}
\email{jiaxin.liu6@student.unsw.edu.au}

\author{Shobhit Sinha}
\authornotemark[1]
\affiliation{%
  \institution{University of Illinois Urbana-Champaign}
  \city{Champaign}
  \country{USA}}
\email{ss194@illinois.edu}

\author{Hao Xue}

\affiliation{%
 \institution{University of New South Wales}
 \city{Sydney}
 \state{NSW}
 \country{Australia}}
\email{hao.xue1@unsw.edu.au}

\author{Flora Salim}
\affiliation{%
  \institution{University of New South Wales}
  \city{Sydney}
  \state{NSW}
  \country{Australia}}
\email{flora.salim@unsw.edu.au}


\renewcommand{\shortauthors}{Chenlu Ju, Jiaxin Liu, Shobhit Sinha, Hao Xue, \& Flora Salim}

\begin{abstract}
This work leverages Large Language Models (LLMs) to simulate human mobility, addressing challenges like high costs and privacy concerns in traditional models. Our hierarchical framework integrates persona generation, activity selection, and destination prediction, using real-world demographic and psychological data to create realistic movement patterns. Both physical models and language models are employed to explore and demonstrate different methodologies for human mobility simulation. By structuring data with summarization and weighted density metrics, the system ensures scalable memory management while retaining actionable insights. Preliminary results indicate that LLM-driven simulations align with observed real-world patterns, offering scalable, interpretable insights for social problems such as urban planning, traffic management, and public health. The framework’s ability to dynamically generate personas and activities enables it to provide adaptable and realistic daily routines. This study demonstrates the transformative potential of LLMs in advancing mobility modeling for societal and urban applications. The source code and interactive demo for our framework are available at \url{https://github.com/cju0/TrajLLM}.
\end{abstract}



\maketitle

\section{Introduction}

A comprehensive understanding of human mobility patterns is essential for the development of sustainable communities. It plays a pivotal role in the control of infectious diseases, in optimizing energy consumption, and in the formulation of urban planning strategies. However, direct utilization of real-world mobility data is often constrained by privacy concerns, as it encompasses sensitive information regarding individual movement patterns and personal behaviors. Furthermore, the collection of comprehensive mobility data is expensive. In this case, simulation of mobility data has emerged as a promising solution. 

Previous approaches to modeling human mobility can be broadly categorized into mechanistic and deep learning models. Mechanistic models leverage stochastic processes to characterize human mobility behaviors, while these models tend to be oversimplified and focus only on certain aspects of mobility behavior, failing to generate comprehensive and nuanced mobility patterns. In contrast, deep learning models are capable of learning and generating human mobility patterns from training data, but often fail to capture the underlying mechanisms driving mobility behavior. Additionally, they suffer from low sample efficiency when learning behavior distributions and tend to generate data that lacks semantic-aware intentions, limiting their interpretability and realism.

The advances in Large Language Models (LLMs) have shown promising capabilities in understanding and generating coherent intentions through techniques like chain-of-thought prompting and role-playing. It has demonstrated remarkable abilities in understanding humans and society, with applications in political science, economics, and other social science studies. In recent years, several models have been proposed to simulate human mobility data using LLMs, such as LLMob \cite{wang2024large} and CoPB \cite{CoPB}. LLMob relies solely on LLMs, utilizing historical data and personal profiles, while CoPB integrates LLMs with a gravity model to simulate mobility data.

Our framework TrajLLM combines the ability of LLMs to generate coherent and interpretable human activities with an innovative physical model for location mapping, minimizing reliance on real-world historical check-in data by leveraging static Point-of-Interest (POI) data. Additionally, we integrate a memory module to ensure agents exhibit consistent and realistic activity patterns over time, reflecting habitual behaviors. 
In comparison to existing solutions, this novel approach not only enables the activity-driven generation of daily mobility trajectories, but also further enhances agent alignment with human behaviors through the consolidation of agents' historical visitation data. Furthermore, the integration of fundamental physical models ensures TrajLLM's simulation efficiency, providing a balance between accuracy and performance.

\section{Framework Architecture}

\begin{figure} [ht]
    \centering
    \includegraphics[width=\linewidth]{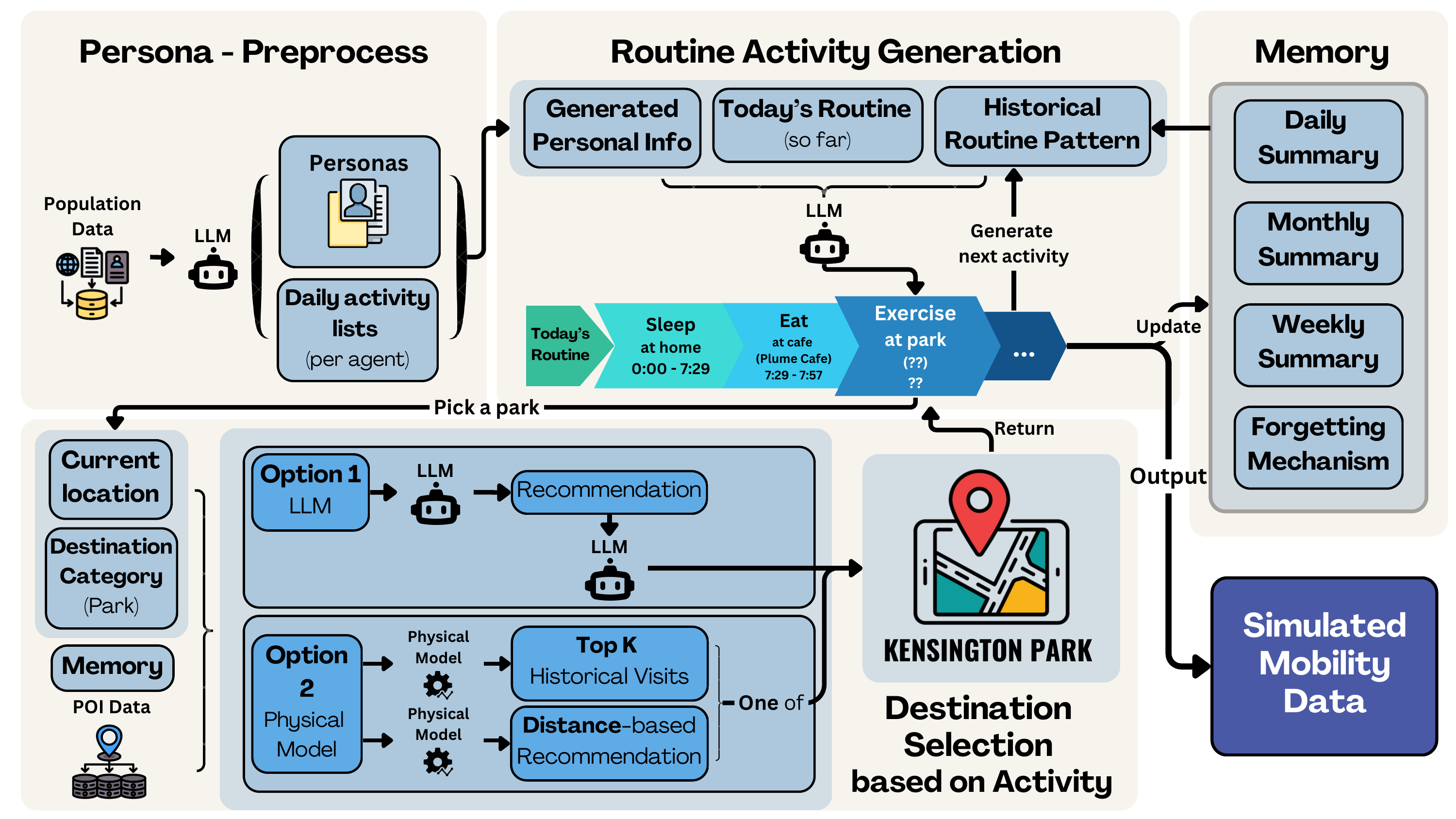}
    \caption{Overall Pipeline of TrajLLM}
    \Description{This is a figure showing the overall pipeline of TrajLLM. It shows the interactions between persona, activity, destination and memory modules during simulation.}
    \label{fig:pipeline}
\end{figure}

To integrate the reasoning abilities of LLMs into mobility simulation and model the activity-driven dynamics of human mobility, our framework consists of four primary modules: persona, activity, destination, and memory.
Figure~\ref{fig:pipeline} illustrates the overall framework pipeline and interactions between the four modules.

To take both time and cost efficiency into account, we adopt LLAMA-3.1-8B-Instruct and GPT-4o-mini as LLM options to check the performance of these two models.

\subsection{Persona}
The first module of TrajLLM focuses on creating agents for mobility simulation. It consists of two phases: generating personas and generating daily activity-location lists.

To effectively generate personas that align with real-world population distribution conditions, population statistics are collected from government websites and prompted into LLMs. The statistics are collected for attributes including age, gender, employment condition, and occupation, which are assumed to be the key attributes that affect the agent's daily activity pattern. Inspired by the personality simulation capability of LLMs~\cite{personaLLM}, personas are also assigned with big five personality traits~\cite{big5}. This allows LLM-empowered agents to simulate mobility pattern differences between people with different personalities. As residences, workplaces (if the agent is an employee), and schools (if the agent is a student) are relatively constant in reality, each persona will have fixed primary locations assigned to them. These locations will serve as key anchors in their daily trajectories, ensuring a realistic representation of routine movements and interactions within the simulated environment.

Once personas are properly generated, each persona is assigned a unique daily activity-location list, which is generated by LLMs with the persona data.
The list details the specific activities the corresponding agent is likely to engage in, and their associated potential location categories. For example, an agent's `meal' activity may have a list of potential location categories: [`Cafe', `Casual Dining', `Home', `Restaurant'], which means one of these categories may be selected as the dining location for the agent.

This module is a pre-processing step before the actual operation of the mobility simulation. Once personas and their corresponding daily activity-location lists are generated, they are utilized as foundational attributes for the agents. These attributes guide the agents' behavior, enabling realistic modeling of human mobility patterns based on diverse profiles and daily routines.

\subsection{Activity}
As illustrated in Figure~\ref{fig:pipeline}, after the persona module is pre-processed, the generated personas are passed into the activity module as the agents' background information. 
For each LLM-empowered agent, the subsequent activity of the day is simulated iteratively based on a combination of persona attributes, potential activity and location categories, the agent's routine up to that point, and historical mobility patterns. This set of information is then used to create a prompt for the LLM, enabling the language model to reason and generate contextually appropriate activity predictions for the persona. The output data for each activity includes the type of activity, the corresponding location category, and its duration.

Once the location category for an agent's activity is determined by the output of the activity module, it is forwarded to the destination module, which selects a specific destination within the given category. After the destination is finalized, the activity module then starts generating the agent's next activity based on contextual factors and predefined rules. This iterative simulation continues until the end of the day (where the day's activities are all completed), at which point the process resets for the next agent.

\subsection{Destination}
In this section, we focus on determining the destinations for the generated daily activities. As illustrated in Figure~\ref{fig:pipeline}, we have designed two destination selection mechanisms: one is driven by LLM and the other utilizes physical models. Both of them take in the current location of the agent (as coordinate), location category for the activity and the agent's historical visiting data from memory.

\subsubsection{LLM-based Destination Selection}
This approach relies entirely on Large Language Models (LLMs) to generate contextually relevant destination recommendations. Historical visiting data is retrieved from the memory module, which stores detailed logs of previous activities and preferences. LLMs process this data to uncover patterns and create recommendations tailored to the persona's goals and behavioural history.

Once recommendations are generated, they serve as input prompts for LLMs to randomize a suitable destination matching a given location category within a specified radius. 
For example, if the activity is "sports and exercise" with a preference for "Gym," the LLM identifies relevant past visits and incorporates them to suggest a nearby gym from the Points of Interests (POIs) dataset.

This method ensures that destination selection is personalized and diverse, balancing historical relevance with real-time randomization. Additionally, the integration of historical activity data ensures that generated recommendations align closely with realistic behavioural patterns, providing a robust foundation for simulated agent movements.

\subsubsection{Physical Model based Destination Selection}
In this module, we consider using mechanistic methods and integrating spatial and historical factors through two primary modules: the Frequency Module and the Spatial Module. The Spatial Module models are based on the spatial interactions between POIs, while the Frequency Module is based on historical data which includes information about agents' preferences and historical patterns. By combining these two aspects, the model ensures that the generated trajectories closely mimic real-world mobility behaviors, achieving both coherence and representativeness in synthetic data generation.

We propose a potential-based spatial model inspired by spatial interaction theory and electromagnetic field principles. The original model is expressed as:
\begin{equation} 
    E(Y_{ij}) = K U_iV_jf(d_{ij})
\end{equation}
where $E(Y_{ij})$is the expected interaction of travel flow $Y_{ij}$ from user current location $i$ to next location $j$, $U_i$ and $V_j$ are attraction factors , $K$ is a scaling coefficient. $f(d_{ij})$ is the distance impedance function and $d_{ij}$ is the distance between the location $i$ and $j$.

In our case, since we only need to predict the next location for a single agent, we use spatial weight $W^s_{j}$ to represents the probability of the user moving from the current location to the next location  $j$. So the scaling coefficient $K$ and $U_i$ are omitted, and our spatial model weight of each POI is defined as:
\begin{equation}
    W^s_j =  \frac{V_j}{f(d_{ij})}
\end{equation}

We chose a truncated power law distribution \cite{gonzalez2008mobility} as the distance impedance function, which is defined as:

\begin{equation}
    f(d) = (d + r_0)^{-\beta}e^{-d/k}
\end{equation}
where $r_0$, $\beta$, and $k$ are parameters estimated from empirical data.

The model incorporates historical visits and personal preferences. To manage the large number of POIs in urban areas, we use the lossy count method to discard low-frequency POIs, conserving memory.  

To ensure interpretable predictions, input frequencies are adjusted via quantile mapping, aligning them with real-world distributions:
\begin{enumerate}
    \item \textbf{Rank normalization:} Standardize input frequencies $f_i$ using the inverse ECDF of check-in data:
    \begin{equation}
    z_i = F_c^{-1}\left(\frac{\text{rank}(f_i)}{|P_c|}\right)
    \end{equation}
    where $F^{-1}_c$ represents the inverse ECDF of the target distribution estimated from check-in data of POI category $c$, $P_c$ is the POI set of category $c$.
    \item \textbf{Distribution mapping:} Map standardized values using a function $\Psi$, resulting in adjusted frequencies:
    \begin{equation}
    f_i' = \Psi(z_i)
    \end{equation}
    \item \textbf{Weight normalization:} Calculate final weights by normalizing \(f_i'\) and introducing a noise factor $\sigma$:
    \begin{equation}
    W^f_i = (1-\sigma) \frac{f_i'}{\sum_{j \in P_c} f_j'} + \sigma \frac{1}{|P_c|}
    \end{equation}
    where $W^f_i$ represent normalized spatial and frequency weights.
\end{enumerate}

Unlike additive methods, experiments show that multiplicative integration better captures the complementary relationship between spatial and frequency weights. To ensuring a realistic balance between convenience and preference, The probability for POI $i$ is computed as:
\begin{equation}
   P_i = \frac{W^s_i \cdot W^f_i}{\sum_{j \in \mathcal{P_c}} W^s_j \cdot W^f_j}
\end{equation}

\subsection{Memory}
The memory module serves as a foundational component of the simulation framework, enabling efficient management of the vast data streams generated by agents. It employs a hierarchical structure, organizing raw daily activities into daily, weekly, and monthly summaries. This progressive summarization captures critical patterns and trends while minimizing redundancy, ensuring the system remains scalable even as interactions grow over time. 

A standout feature of the memory module is the use of weighted information density metrics to evaluate memory significance. Drawing inspiration from studies such as Holmes and Rahe’s (1967)~\cite{social} Social Readjustment Rating Scale and Tiwari and Deshpande’s (2023)~\cite{urban} urban stress evaluation, the system assigns weights to categories like events, entities, actions, and attributes to reflect their relative importance. While these references provided a conceptual basis for assigning weights, the actual numerical values were determined independently during development. Events and entities, often pivotal in contextualizing activities, are weighted higher to prioritize retention of meaningful details. 

The module calculates an importance score for each memory by combining weighted information density with recency and access frequency, normalized using a sigmoid function to produce a balanced, interpretable measure of relevance. This approach ensures that high-density memories remain prioritized while accommodating temporal factors like recentness and usage patterns. Through this scoring, the system identifies which memories to retain and which to prune, striking a balance between utility and storage efficiency. Extensive iterative testing helped establish a robust threshold score to dynamically prune memories, retaining those with actionable insights while discarding less critical ones. This ensures efficient and relevant memory management.

By balancing scalability with depth, the memory module adapts dynamically to agent needs, supporting realistic and adaptive simulations. Future improvements will refine the weighting system and incorporate personalized metrics to better align memory management with individual agent personas and their unique interactions.

\section{Demonstration}

Our solution integrates back-end processing and a dynamic front-end to create an intuitive simulation framework. It consists of two key components: (1) a Flask-based back-end server for handling simulation logic and managing agent activity data, and (2) a front-end web application built with HTML, CSS, JavaScript, and the Leaflet.js library for visualization.

The back-end server provides RESTful API endpoints, such as the \texttt{/start\_day} endpoint, which processes input parameters like the number of agents and start time to generate a JSON response with detailed agent activities and coordinates.
Its modular architecture enables easy integration with new data sources and generative models, ensuring scalability and adaptability.

The front-end application offers a user-friendly interface for real-time interaction. Users can adjust parameters via sliders and input fields, visualize agent movements on a map, and access live activity logs with toggle-able controls. This combination ensures usability and interactivity, making it ideal for engaging demonstrations. The front-end HTML file is available in our GitHub repository for users to explore and interact with independently.

To ensure accessibility and usability during demonstrations, the framework is deployed on a web server using Flask. This setup allows conference attendees to interact with the simulation directly on their devices or via a dedicated terminal provided at the venue.

\begin{figure}[ht]
    \centering
    \includegraphics[width=\linewidth]{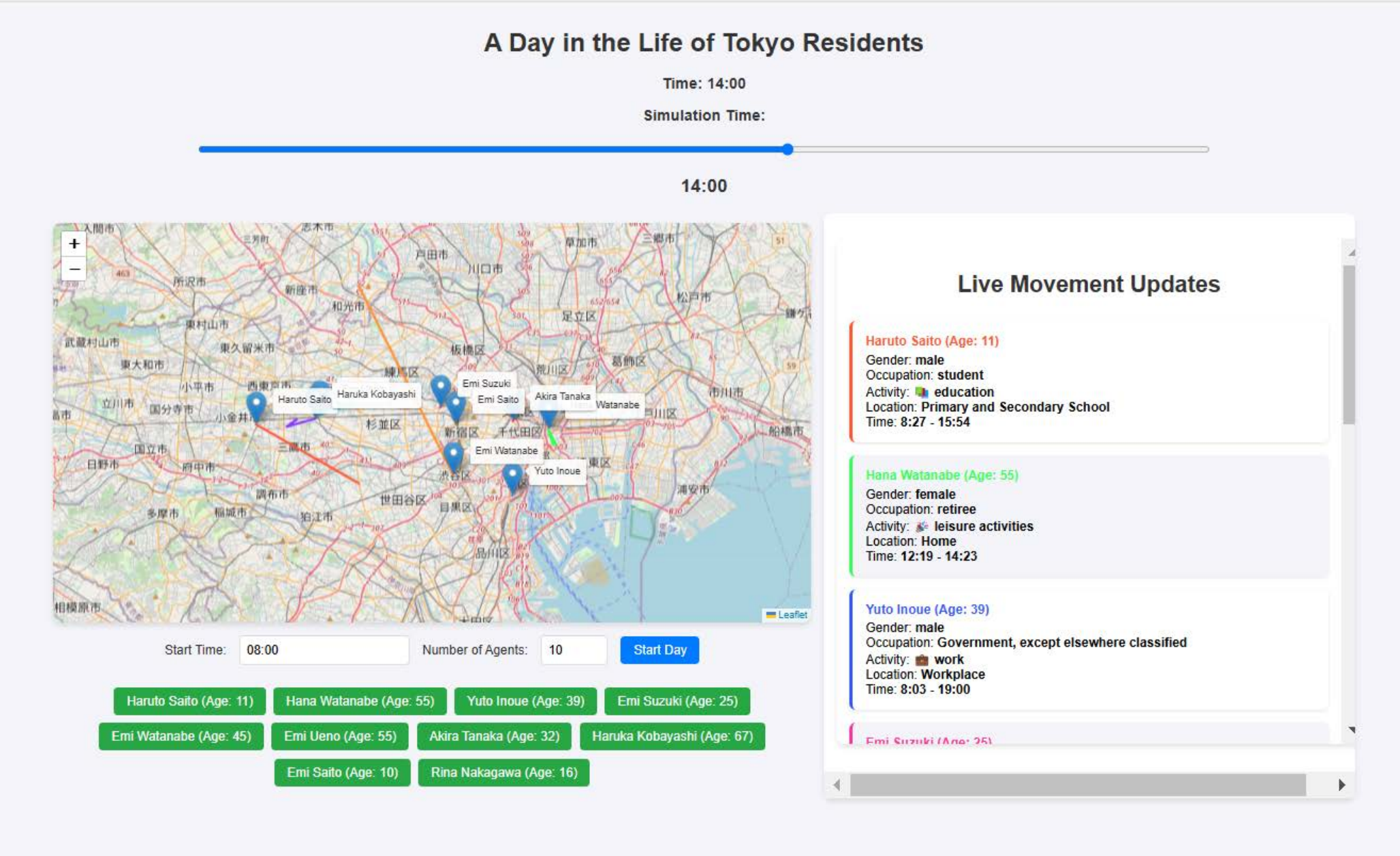}
    \caption{Real-Time Simulation of Daily Activities in Tokyo}
    \Description{This is a figure showing the front-end demonstration for the simulation of one-day routines of 10 agents in Tokyo. Their movement for daily activities are represented on the map, with textual details of the activities.}
    \label{fig:demo}
\end{figure}

For the purpose of demonstration, we leverage the public check-ins dataset of Tokyo as the underlying POIs dataset for simulation (note that the proposed simulation platform supports other cities as well). Based on this, the personas for the agents are created using population statistics collected from local government
website\footnote{\url{https://www.stat.go.jp/english/data/index.html}}.
As illustrated in Figure~\ref{fig:demo}, the visualization presents a sample of one-day trajectories for ten agents in Tokyo.
This interface provides an intuitive layout with a map view for agent trajectories, sliders, and input fields for simulation parameters. The "Live Movement Updates" panel displays agent-specific details such as gender, occupation, activity, and location, highlighting the system's scalability, modularity, and ease of interaction.
This makes it an effective tool for showcasing dynamic simulations in academic and professional environments.

\section{Conclusion}

In this paper, we propose an LLM-empowered agent-based modular mobility generation framework, TrajLLM. It utilizes iterative activity simulation and hybrid models destination selection mechanisms to allow memory-guided LLM agents to simulate effective human activity trajectories. Key contributions include the development of a modular framework that seamlessly integrates human behavioral and spatial-temporal dynamics into mobility simulation, significantly enhancing the realism and adaptability of generated long-term human movement patterns, while reducing the reliance on real-world data. In addition, our interactive web interface allows users to flexibly configure simulation parameters to visualize the daily trajectories of different numbers of agents at different times.

The limitations of LLMs in fully capturing the complexity of human behavior are undeniable, but our activity-driven trajectory simulation mechanism enhances overall accuracy. However, the potential biases embedded within the data sets present a persistent challenge that requires further mitigation. Furthermore, future research will focus on incorporating interactions between agents to refine the simulation's realism and generalizability.

\section{Ethical Use of Data}

All datasets used in this research are composed of publicly available data obtained from credible sources. These datasets do not contain any personally identifiable information (PII), ensuring compliance with ethical standards for data usage. This study ensures that the data utilized is properly sourced, transparent, and used responsibly, with no potential for harm or privacy violations. 

TrajLLM-generated data should be indicated as synthetic to avoid misleading conclusions in usage and to ensure accountability and transparency.

\begin{acks}
We would like to thank the support of the National Computational Infrastructure, the ARC Center of Excellence for Automated Decision Making and Society (CE200100005), and OpenAI’s Researcher Access Program for providing API access to GPT models.
\end{acks}

\bibliographystyle{ACM-Reference-Format}
\balance
\bibliography{reference}


\begin{thebibliography}{7}


\ifx \showCODEN    \undefined \def \showCODEN     #1{\unskip}     \fi
\ifx \showDOI      \undefined \def \showDOI       #1{#1}\fi
\ifx \showISBNx    \undefined \def \showISBNx     #1{\unskip}     \fi
\ifx \showISBNxiii \undefined \def \showISBNxiii  #1{\unskip}     \fi
\ifx \showISSN     \undefined \def \showISSN      #1{\unskip}     \fi
\ifx \showLCCN     \undefined \def \showLCCN      #1{\unskip}     \fi
\ifx \shownote     \undefined \def \shownote      #1{#1}          \fi
\ifx \showarticletitle \undefined \def \showarticletitle #1{#1}   \fi
\ifx \showURL      \undefined \def \showURL       {\relax}        \fi
\providecommand\bibfield[2]{#2}
\providecommand\bibinfo[2]{#2}
\providecommand\natexlab[1]{#1}
\providecommand\showeprint[2][]{arXiv:#2}

\bibitem[González et~al\mbox{.}(2008)]%
        {gonzalez2008mobility}
\bibfield{author}{\bibinfo{person}{Marta~C. González}, \bibinfo{person}{César~A. Hidalgo}, {and} \bibinfo{person}{Albert-László Barabási}.} \bibinfo{year}{2008}\natexlab{}.
\newblock \showarticletitle{Understanding individual human mobility patterns}.
\newblock \bibinfo{journal}{\emph{Nature}} \bibinfo{volume}{453}, \bibinfo{number}{7196} (\bibinfo{year}{2008}), \bibinfo{pages}{779--782}.
\newblock


\bibitem[Holmes and Rahe(1967)]%
        {social}
\bibfield{author}{\bibinfo{person}{Thomas~H. Holmes} {and} \bibinfo{person}{Richard~H. Rahe}.} \bibinfo{year}{1967}\natexlab{}.
\newblock \showarticletitle{The Social Readjustment Rating Scale}.
\newblock \bibinfo{journal}{\emph{Journal of Psychosomatic Research}} \bibinfo{volume}{11}, \bibinfo{number}{2} (\bibinfo{year}{1967}), \bibinfo{pages}{213--218}.
\newblock


\bibitem[Jiang et~al\mbox{.}(2024)]%
        {personaLLM}
\bibfield{author}{\bibinfo{person}{Hang Jiang}, \bibinfo{person}{Xiajie Zhang}, \bibinfo{person}{Xubo Cao}, \bibinfo{person}{Cynthia Breazeal}, \bibinfo{person}{Deb Roy}, {and} \bibinfo{person}{Jad Kabbara}.} \bibinfo{year}{2024}\natexlab{}.
\newblock \showarticletitle{PersonaLLM: Investigating the Ability of Large Language Models to Express Personality Traits}.
\newblock  (\bibinfo{date}{April} \bibinfo{year}{2024}).
\newblock


\bibitem[Park et~al\mbox{.}(1990)]%
        {big5}
\bibfield{author}{\bibinfo{person}{Joon~Sung Park}, \bibinfo{person}{Joseph~C. O'Brien}, \bibinfo{person}{Carrie~J. Cai}, \bibinfo{person}{Meredith~Ringel Morris}, \bibinfo{person}{Percy Liang}, {and} \bibinfo{person}{Michael~S. Bernstein}.} \bibinfo{year}{1990}\natexlab{}.
\newblock \showarticletitle{Generative Agents: Interactive Simulacra of Human Behavior}.
\newblock \bibinfo{journal}{\emph{Journal of Personality and Social Psychology}} \bibinfo{volume}{59}, \bibinfo{number}{6} (\bibinfo{date}{December} \bibinfo{year}{1990}), \bibinfo{pages}{1216--29}.
\newblock


\bibitem[Shao et~al\mbox{.}(2024)]%
        {CoPB}
\bibfield{author}{\bibinfo{person}{Chenyang Shao}, \bibinfo{person}{Fengli Xu}, \bibinfo{person}{Bingbing Fan}, \bibinfo{person}{Jingtao Ding}, \bibinfo{person}{Yuan Yuan}, \bibinfo{person}{Meng Wang}, {and} \bibinfo{person}{Yong Li}.} \bibinfo{year}{2024}\natexlab{}.
\newblock \showarticletitle{Chain-of-Planned-Behaviour Workflow Elicits Few-Shot Mobility Generation in LLMs}.
\newblock \bibinfo{journal}{\emph{arXiv preprint arXiv:2402.09836}} (\bibinfo{year}{2024}).
\newblock


\bibitem[Tiwari and Deshpande(2020)]%
        {urban}
\bibfield{author}{\bibinfo{person}{Sayali~C. Tiwari} {and} \bibinfo{person}{Swati~R. Deshpande}.} \bibinfo{year}{2020}\natexlab{}.
\newblock \showarticletitle{A study to assess the effect of stressful life events on psychological distress levels of participants living in an urban area}.
\newblock \bibinfo{journal}{\emph{Journal of Family Medicine and Primary Care}} \bibinfo{volume}{9}, \bibinfo{number}{6} (\bibinfo{date}{June} \bibinfo{year}{2020}), \bibinfo{pages}{2730--2735}.
\newblock
\urldef\tempurl%
\url{https://doi.org/10.4103/jfmpc.jfmpc_96_20}
\showDOI{\tempurl}


\bibitem[Wang et~al\mbox{.}(2024)]%
        {wang2024large}
\bibfield{author}{\bibinfo{person}{Jiawei Wang}, \bibinfo{person}{Renhe Jiang}, \bibinfo{person}{Chuang Yang}, \bibinfo{person}{Zengqing Wu}, \bibinfo{person}{Makoto Onizuka}, \bibinfo{person}{Ryosuke Shibasaki}, \bibinfo{person}{Noboru Koshizuka}, {and} \bibinfo{person}{Chuan Xiao}.} \bibinfo{year}{2024}\natexlab{}.
\newblock \showarticletitle{Large language models as urban residents: An llm agent framework for personal mobility generation}.
\newblock \bibinfo{journal}{\emph{arXiv preprint arXiv:2402.14744}} (\bibinfo{year}{2024}).
\newblock


\end{thebibliography}

\end{document}